\documentclass[conference]{IEEEtran}
\IEEEoverridecommandlockouts
\usepackage{cite}
\usepackage{amsmath,amssymb,amsfonts}
\usepackage{algorithmic}
\usepackage{graphicx}
\usepackage{textcomp}
\usepackage{xcolor}
\def\BibTeX{{\rm B\kern-.05em{\sc i\kern-.025em b}\kern-.08em
    T\kern-.1667em\lower.7ex\hbox{E}\kern-.125emX}}

\begin{document}

\title{Real-Time Fruit Recognition and Grasping Estimation for Autonomous Apple Harvesting}

\author{\IEEEauthorblockN{1\textsuperscript{st} Hanwen Kang}
\IEEEauthorblockA{\textit{Monash University} \\
\textit{Department of Mechanical and Aerospace Engineering}\\
Melbourne, Australia \\
hanwen.kang@monash.edu}
\and
\IEEEauthorblockN{2\textsuperscript{nd} Chao Chen}
\IEEEauthorblockA{\textit{Monash University} \\
\textit{Department of Mechanical and Aerospace Engineering}\\
Melbourne, Australia \\
chao.chen@monash.edu}
}
\maketitle

\begin{abstract}
	In this paper, a fully neural network based visual perception framework for autonomous apple harvesting is proposed. The proposed framework includes a multi-function neural network for fruit recognition and a Pointnet network to determine the proper grasp pose to guide the robotic execution. Fruit recognition takes raw input of RGB images from the RGB-D camera to perform fruit detection and instance segmentation. The Pointnet grasping estimation takes the point cloud of each fruit as input, and predict the grasp pose for each fruit as output. The proposed framework is validated by using RGB-D images collected from laboratory and orchard environments, robotic grasping experiments in a controlled environment are also included. Experimental results shows that the proposed framework can accurately localise fruits and estimate the grasp pose for robotic grasping. 
\end{abstract}


\section{Introduction}
Autonomous harvesting plays a significant role in the recent development of the agricultural industry \cite{2019harvesting_robot}. Vision is one of the essential tasks in autonomous harvesting, as it can detect and localise the crop, and guide the robotic arm to perform detachment \cite{2016DLW_1}. Vision tasks in orchard environments are challenging as there are many factors influencing the performance of the system, such as variances in illumination, appearance, and occlusion between crop and other items within the environment. Meanwhile, occlusion between fruits and other items can also decrease the success rate of autonomous harvesting \cite{2014harvesting_robot}. In order to increase the efficiency of harvesting, the vision system should be capable of guiding the robotic arm to detach the crop from a proper approach pose. Overall, an efficient vision algorithm which can robustly perform crop recognition and grasp pose estimation is the key to the success of autonomous harvesting \cite{2016review_visionbased_agr}.

In this work, a fully deep-learning based vision algorithm which can perform real-time fruit recognition and grasping estimation for autonomous apple harvesting by using sensory data from the RGB-D camera is proposed. The proposed method includes two function blocks: fruit recognition and grasping estimation. Fruit recognition applies a one-stage multi-task neural network to perform fruit detection and instance segmentation on colour images. Grasp pose estimation processes the information from the fruit recognition together with depth information to estimate the proper grasp pose for each fruit by using the Pointnet. The following highlights are presented in the paper:
\begin{itemize}
	\item Applying a multi-task neural network to perform fruit detection and instance segmentation on input colour images from RGB-D camera.
	\item Proposing a modified Pointnet-based network to perform fruit modelling and grasp pose estimation by using point clouds from RGB-D camera.
	\item Realising and combining the aforementioned two features to guide the robot to perform autonomous harvesting.
\end{itemize}
The rest of the paper is organised as follows. Section \ref{section:review} reviews the related works on fruit recognition and grasp pose estimation. Section \ref{section:method} introduces the methods of the proposed vision processing algorithm. The experimental setup and results are included in Section \ref{section:experiment}. In Section \ref{section:conclusion}, conclusion and future works are presented.

\section{Literature Review}\label{section:review}
\subsection{Fruit Recognition}
Fruit recognition is an essential task in the autonomous agricultural applications \cite{2012TMW_review2}. There are many methods which have been studied in decades, including the traditional method \cite{2019traditional_1, 2019traditional_2, 2019traditional_3} and deep-learning based method. Traditional method applies hand-crafted feature descriptors to describe the appearances of objects within images, and uses machine-learning algorithm to perform classification, detection, or segmentation by using extracted feature descriptors \cite{2012traditional_machine_learning}. The performance of the traditional method is limited by the express ability of the feature descriptor, which required to be adjusted before applying in different conditions \cite{2015deepreview}. Deep-learning based method applies deep convolution neural network to perform automatic image feature extraction, which has shown the good performance and generalisation in many core tasks of the computer vision \cite{2018deep_survey}. Deep-learning based detection method can be divided into two classes: two-stage detection and one-stage detection \cite{2019object_detection_review}. Two-stage detection applies a Region Proposal Network (RPN) to search the Region of Interest (RoI) from the image, and a classification branch is applied to perform bounding box regression and classification \cite{2015fast-rcnn, 2015faster-rcnn}. One-stage detection combines the RPN and classification into a single architecture, which speeds up the processing of the images \cite{2018yolov3, 2016ssd}. Both two-stage detection and one-stage detection have been widely studied in autonomous harvesting \cite{2020kangfast}. Bargoti and Underwood \cite{ 2017DLW_2} applied Faster Region Convolution Neural Network (Faster-RCNN) to perform multi-class fruit detection in orchard environments. Yu et al. \cite{2019DLW_3} applied Mask-RCNN \cite{2017mask-rcnn} to perform strawberry detection and instance segmentation in the non-structural environment. Liu et al. \cite{2019kiwi} applied a modified Faster-RCNN on kiwifruit detection by combining the information from RGB and NIR images, an accurate detection performance was reported in this work. Tian et al. \cite{2019DLW_4} applied an improved Dense-YOLO to perform monitoring of apple growth in different stages. Koirala et al. \cite{2019DLW_5} applied a light-weight YOLO-V2 model which named as 'Mongo-YOLO' to perform fruit load estimation. Kang and Chen \cite{2019DLW_6, 2020DLW_7} develop a multi-task network based on YOLO, which combines the semantic, instance segmentation, and detection in a one-stage network. To efficiently perform the robotic harvesting, the grasping estimation which can guide the accurate robotic harvesting is also required \cite{2019guava}. The aforementioned studies only applied detection network to perform fruit recognition while lack the ability of the grasping estimation.

\subsection{grasping estimation}
Grasp pose estimation is one of the key techniques in the robotic grasp \cite{2012grasp_1}. Similar to the methods developed for fruit recognition, the grasp pose estimation methods can be divided into two categories: traditional analytical approaches and deep-learning based approaches \cite{2018grasp_4}. Traditional analytical approaches extract feature/key points from the point clouds and then perform matching between sensory data and template from the database to estimate the object pose \cite{2012grasp_3}. The pre-defined grasp pose can be applied in this condition. For the unknown objects, some assumption can be made, such as grasp the object along the principle axis \cite{2012grasp_1}. The performance of the traditional analytical approaches is limited when being performed in the real world, the noise or partial point cloud can severely influence the accuracy of the estimation \cite{2017grasp_2}. In the following development, deep-learning based methods recast the grasp pose estimation as an object detection task, which can directly produce grasp pose from the images \cite{2015grasp_5}. Recently, with the development of the deep-learning architecture for 3D point cloud processing \cite{2017pointnet, 2017pointnet++}, some studies focus on performing grasp pose estimation by using the 3D point clouds. These methods apply convolution neural network architectures to process the 3D point clouds and estimate the grasp pose to guide the grasping, such as Grasp Pose Detection (GPD) \cite{2016grasp_6} and Pointnet GPD \cite{2019grasp_7}, which showed accurate performance in the specific conditions. In the robotic harvesting case, Lehnert et al. \cite{2016ROBOT_1} modeled the sweep pepper as the super-ellipsoid and estimated the grasp pose by performing shape matching between the super-ellipsoid and fruit. In their following work \cite{2017robot_2}, surface normal orientation of fruits were applied as grasp candidates and ranked by the an utility function, which is time consuming and not robust to the outdoor environments. Some other studies \cite{2015robot_4, 2016robot_3, 2019robot_5} performed the grasping by translating towards the fruits, which can not secure the success rate of harvesting in unstructured environments. The aforementioned studies are limited to be applied in the specific conditions or not accurate and robust to the orchard environments. In this study, a Pointnet based grasping estimation is proposed to perform fruit modelling and grasp pose estimation by combining with the fruit recognition, which shows the accurate and robust performance in the experiments.

\section{Methods and Materials}\label{section:method}
\subsection{System Design}
\begin{figure}[ht]
	\centering
	\includegraphics[width=.47\textwidth]{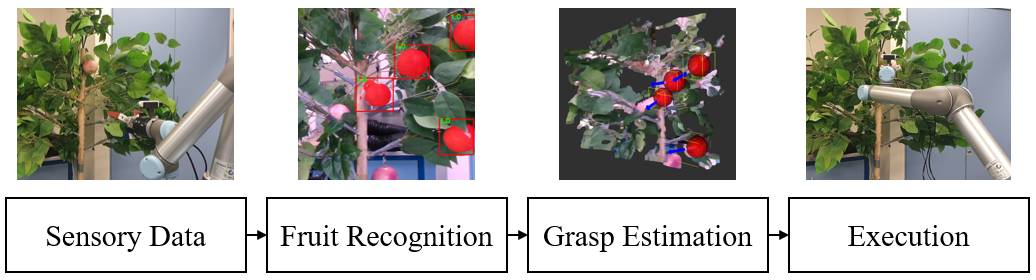}
	\caption{Two-stage vision perception and grasping estimation for autonomous harvesting.}
	\label{fig:workflow}
\end{figure}
The proposed method include two-stages: fruit recognition and grasp pose estimation. The workflow of the proposed vision processing algorithm is shown in Figure \ref{fig:workflow}. In the first step, the fruit recognition block performs fruit detection and segmentation on input RGB images from the RGB-D camera. The outputs of the fruit recognition are projected to the depth images, and the point clouds of each detected fruit are extracted and sent to the grasp pose estimation block for further processing. In the second step, the Pointnet architecture is applied to estimate the geometry and grasp pose of fruits by using the point clouds from the previous steps. The method of the fruit recognition block and grasp pose estimation are presented in Section \ref{section:method_recognition} and \ref{section:method_estimation}, respectively. The implementation details of the proposed method are introduced in Section \ref{section:method_implementation}.

\subsection{Fruit Recognition}\label{section:method_recognition}
\subsubsection{Network Architecture}
\begin{figure}[htbp]
	\centerline{\includegraphics[width=.47\textwidth]{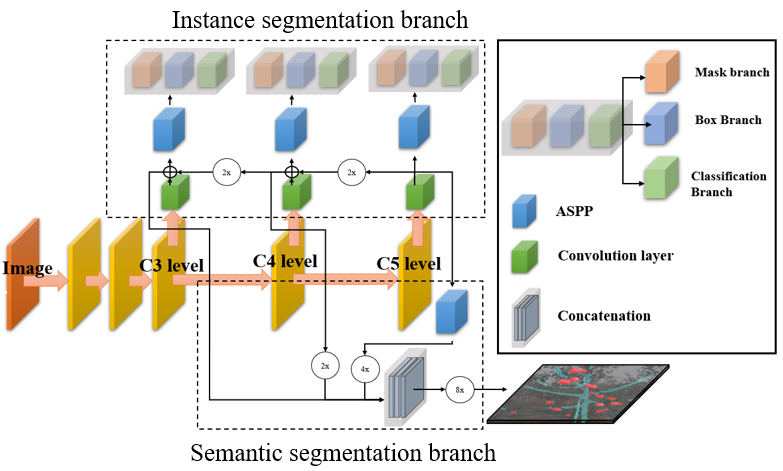}}
	\caption{Network architecture of the Dasnet \cite{2020dasnt_v2}, Dasnet is a one-stage detection network which combines detection, instance segmentation and semantic segmentation.}
	\label{fig:dasnet}
\end{figure}
A one-stage neural network Dasnet \cite{2020dasnt_v2} is applied to perform fruit detection and instance segmentation tasks. Dasnet applies a 50 layers residual network (resnet-50) \cite{2016resnet} as the backbone to extract features from the input image. A three levels Feature Pyramid Network (FPN) is used to fuse feature maps from the C3, C4, and C5 level of the backbone (as shown in Figure \ref{fig:dasnet}). That is, the feature maps from the higher level are fused into the feature maps from the lower level since feature maps in higher level include more semantic information which can increase the classification accuracy \cite{2014ZFNET}. 

\begin{figure}[htbp]
	\centerline{\includegraphics[width=.4\textwidth]{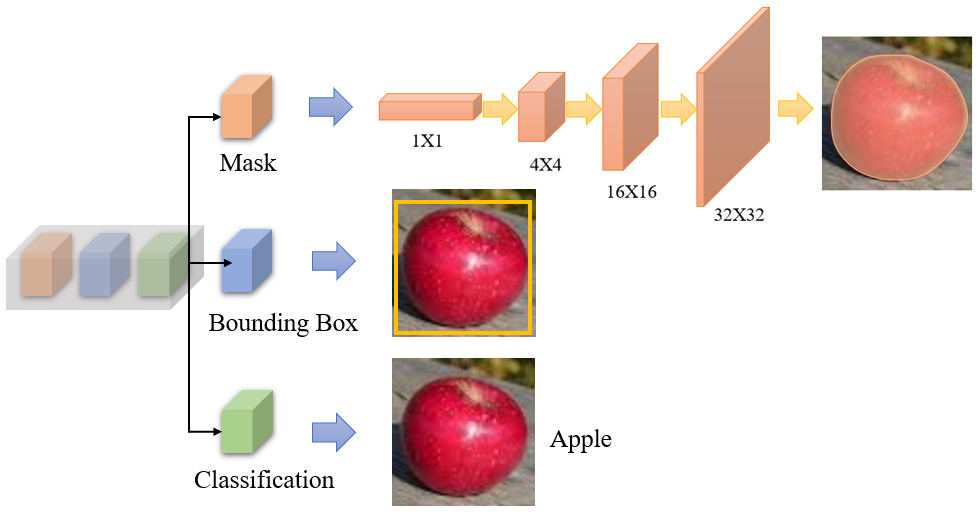}}
	\caption{Architecture of the instance segmentation branch, which can perform instance segmentation, bounding box regression, and classification.}
	\label{fig:branch}
\end{figure}
On each level of the FPN, an instance segmentation (includes detection and instance segmentation) branch is applied, as shown in Figure \ref{fig:branch}. Before the instance segmentation branch, an Atrous Spatial Pyramid Pooling (ASPP) \cite{2017ASPP} is used to process the multi-scale features within the feature maps. ASPP applies dilation convolution with different rates (e.g.1, 2, 4 in this work) to process the feature, which can process the features of different scale separately. The instance segmentation branch includes three sub-branches, which are mask generation branch, bounding box branch, and classification branch. Mask generation branch follows the architecture design proposed in Single Pixel Reconstruction Network (SPRNet) \cite{2019SPRNet}, which can predict a binary mask for objects from a single pixel within the feature maps. Bounding box branch includes the prediction on confidence score and the bounding box shape. We apply one anchor bounding box on each level of FPN (size of anchor box of instance segmentation branch on C3, C4 and C5 level are 32 x 32 (pixels), 80 x 80, and 160 x 160, respectively.). Classification branch predicts the class of the object within the bounding box. The combined outputs from the instance segmentation branch form the results of the fruit recognition on colour images. Dasnet also has a semantic segmentation branch for environment semantic modelling, which is not applied in this research.

\subsubsection{Network Training}
More than 1000 images are collected from apple orchards located in Qingdao, China and Melbourne, Australia. Types of apples, includes Fuji, Gala, Pink Lady, and so on. The images are labelled by using LabelImage tool from Github \cite{LabelImg}. We applied 600 images as the training set, 100 images as the validation set, and 400 images as the test set. We introduce multiple image augmentations in the network training, including random crop, random scaling (0.8-1.2), flip (horizontal only), random rotation ($\pm10^{\circ}$), randomly adjust on saturation (0.8-1.2) and brightness (0.8-1.2) in HSV colour space. We apply focal loss \cite{2017focal} in the training and Adam-optimiser is used to optimise the network parameters. The learning rate and decay rate of the optimiser are 0.001 and 0.75 per epoch. We train the instance segmentation branch for 100 epochs and train the whole network for another 50 epochs.

\subsubsection{Post Processing}
The results of the fruit recognition are projected into the depth image. That is, the mask region of each apple on depth image is extracted. Then, the 3D position of each point in the point clouds of each apple is calculated and obtained. The generated point clouds are the visible part of the apple from the current view-angle of the RGB-D camera. These point clouds are further processed by grasp pose estimation block to estimate the grasp pose, which is introduced in the following section.

\subsection{grasping estimation}\label{section:method_estimation}
\subsubsection{Grasp Planning}
\begin{figure}[htbp]
	\centerline{\includegraphics[width=.47\textwidth]{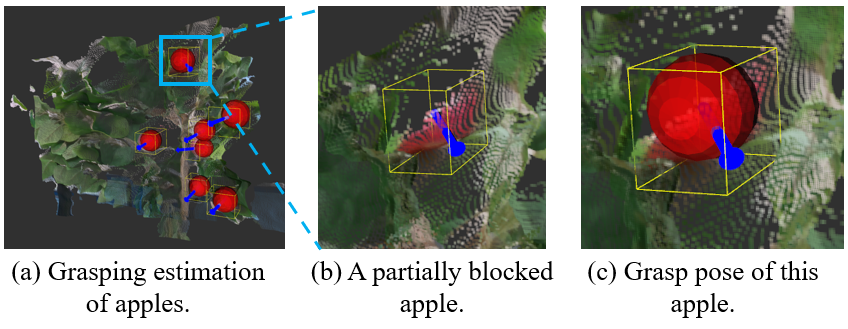}}
	\caption{Our method select orientation from the fruit centre to visible part centre as grasp pose.}
	\label{fig:grasp_est}
\end{figure}

Since most of the apples are presented in sphere or ellipsoid, we modelling the apple as sphere shape for simplified expression. In the natural environments, apples can be blocked by branches or other items within the environments from the view-angle of the RGB-D camera. Therefore, the visible part of the apple from the current view-angle of the RGB-D camera indicates the potential grasp pose, which is proper for the robotic arm to pick the fruit. Unlike GPD \cite{2016grasp_6} or Pointnet GPD \cite{2019grasp_7} which generates multiple grasp candidates and uses the network to determine the best grasp pose , we formulate the grasp pose estimation as an object pose estimation task which is similar to the Frustum PointNets \cite{2018frustum}. We select the centre of the visible part and orientation from the centre of the apple to this centre as the position and orientation of the grasp pose (as shown in Figure \ref{fig:grasp_est}). The Pointnet takes 1-viewed point cloud of each fruit as input and estimates the grasp pose for the robotic arm to perform detachment.

\subsubsection{Grasp Representation}\label{section:method_estimation_representation}
\begin{figure}[htbp]
	\centerline{\includegraphics[width=.47\textwidth]{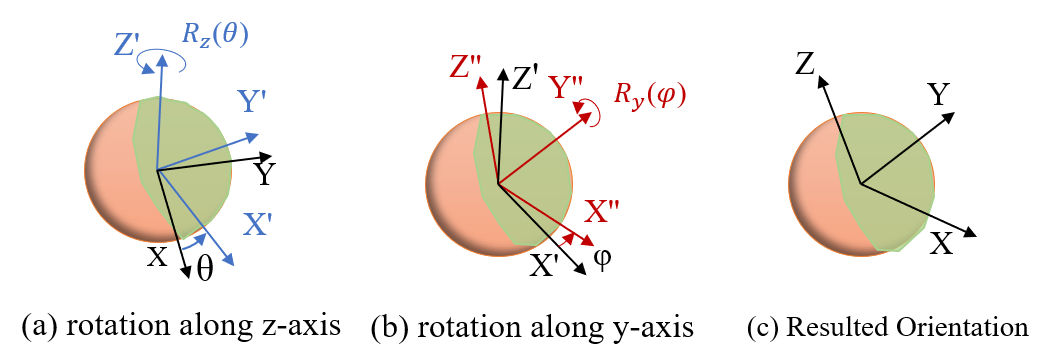}}
	\caption{Euler-ZYX angle is applied to represent the orientation of the grasp pose.}
	\label{fig:EULAR}
\end{figure}

The pose of an object in 3D space has 6 Degree of Freedom (DoF), includes three positions (x, y, and z), and three rotations ($\theta$, $\phi$, and $\omega$, along Z-axis, Y-axis, and X-axis, respectively). We apply Euler-ZYX angle to represent the orientation of the grasp pose, as shown in Figure \ref{fig:EULAR}. The value of $\omega$ is set as zero since we can always assume that fruit will not rotate along its X-axis (since apples are presented in a spherical shape). The grasp pose (GP) of an apple can be formulated as follow:
\begin{equation}{T_{GP}=
	\left[\begin{array}{cccc}
	\cos\theta \cos\phi & -\sin\theta & \cos\theta \sin\phi &x\\
	\sin\theta \cos\phi & \cos\theta & \sin\theta \sin\phi &y\\
	-\sin\phi & 0 & \cos\phi &z\\
	0 & 0 & 0 & 1
	\end{array} 
	\right]}
\end{equation} 
Therefore, a parameter list [x, y, z, $\theta$, $\phi$] is used to represent the grasp pose of the fruit.

\subsubsection{Data Annotation}\label{section:method_estimation_annotation}
Grasp pose block use point clouds as input and predicts the 3D Oriented Bounding Box (3D-OBB) (oriented in grasp orientation) for each fruit. Each 3D-OBB includes six parameters, which are $x$, $y$, $z$, $r$, $\theta$, $\phi$. The position ($x$, $y$, $z$) represents the offsets on X-, Y-, Z-axis from the centre of point clouds to the centre of the apple, respectively. The parameter $r$ represents the radius of the apple, as the apples is modelled as sphere. The length, width, and height can be derivated by radius. $\theta$ and $\phi$ represent the grasp pose of the fruit, as described in Section \ref{section:method_estimation_representation}.  

Since the values of the parameters $x$, $y$, $z$, and $r$ may have large variances when dealing with prediction in different situations, a scale parameters $S$ is introduced. We apply $S$ to represent the mean scale  (radius) of the apple, which equals 30 (cm) in our case. The parameters $x$, $y$, $z$, and $r$ are divided by $S$ to obtain the united offset and radius ($x_{u}$, $y_{u}$, $z_{u}$, $r_{u}$). After remapping, the range of the $x_{u}$, $y_{u}$, $z_{u}$ is reduced to [-$\infty$, $\infty$], and the range of $r_{u}$ are in [0, $\infty$]. To keep the grasp pose in the range of motion of the robotic arm, the $\theta$ and $\phi$ are limited in the range of [$-\frac{1}{4}\pi$, $\frac{1}{4}\pi$]. We divide the $\theta$ and $\phi$ by $\frac{1}{4}\pi$ to map the range of grasp pose into the range of [-1,1]. The united $\theta$ and $\phi$ are denoted as $\theta_{u}$ and $\phi_{u}$. In total, we have six united parameters to predict the 3D-OBB for each fruit, which are [$x_{u}$, $y_{u}$, $z_{u}$, $r_{u}$, $\theta_{u}$, $\phi_{u}$]. Among these parameters, [$x_{u}$, $y_{u}$, $z_{u}$, $\theta_{u}$, $\phi_{u}$] represent the grasp pose of the fruit, $r_{u}$ controls the shape of 3D-OBB.

\subsubsection{Pointnet Architecture}
\begin{figure}[htbp]
	\centerline{\includegraphics[width=.4\textwidth]{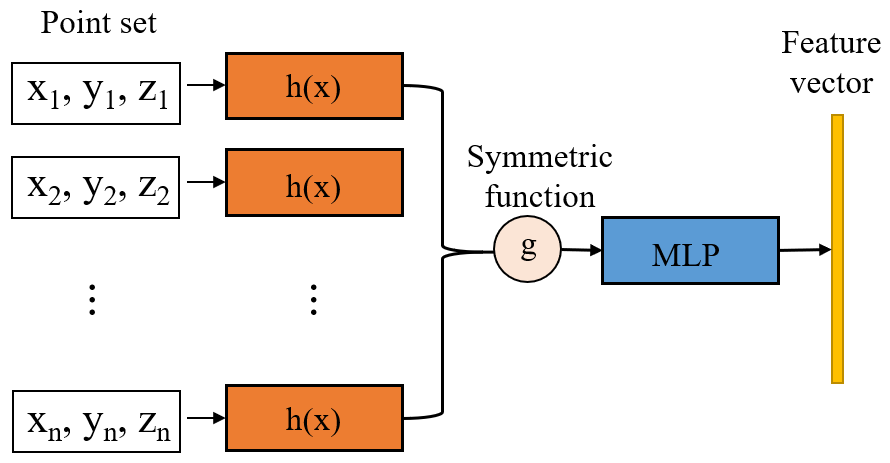}}
	\caption{Pointnet applies symmetric function to extract features from the unordered point cloud.}
	\label{fig:pn_principle}
\end{figure}
\begin{figure}[htbp]
	\centerline{\includegraphics[width=.47\textwidth]{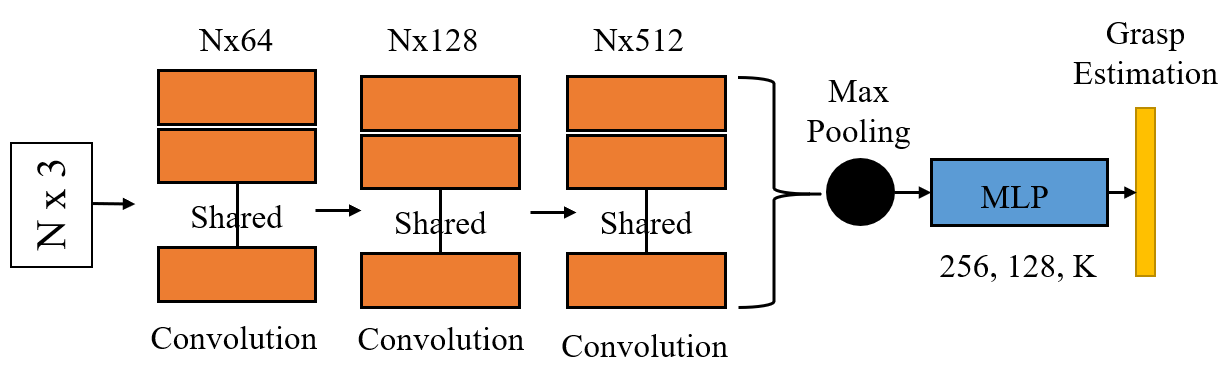}}
	\caption{Network architecture of the Pointnet applied in grasping estimation.}
	\label{fig:pn_architecture}
\end{figure}

Pointnet \cite{2017pointnet} is a deep neural network architecture which can perform classification, segmentation, or other tasks on point clouds. Pointnet can use raw point clouds of the object as input and does not requires any pre-processing. The architecture of the Pointnet is shown in Figure \ref{fig:pn_principle} and \ref{fig:pn_architecture}. Pointnet uses an n x 3 (n is the number of points) unordered point clouds as input. Firstly, Pointnet applies convolution operations to extract a multiple dimensional feature vector on each point. Then, a symmetric function is used to extract the features of the point clouds on each dimension of the feature vector. 
\begin{equation}\label{eq:symmetric}
f(x_{1}, x_{2}, ... ,x_{n})=g(h(x_{1}), h(x_{2}), ... , h(x_{2}))
\end{equation}
In Eq. \ref{eq:symmetric}, $g$ is a symmetric function and $f$ is the extracted features from the set. Pointnet applies max-pooling as the symmetric function. In this manner, Pointnet can learn numbers of features from point set and invariant to input permutation. The generated feature vectors are further processed by Multi-Layer Perception (MLP) (fully-connected layer in Pointnet), to perform classification of the input point clouds. Batch-norm layer is applied after each convolution layer or fully-connection layer. Drop-out is applied in the fully-connected layer during the training.

In this work, the output of the Pointnet is changed to the 3D-OBB prediction, which includes prediction on six parameters [$x_{u}$, $y_{u}$, $z_{u}$, $r_{u}$, $\theta_{u}$, $\phi_{u}$]. The range of the parameters $x_{u}$, $y_{u}$, and $z_{u}$ are in [-$\infty$, $\infty$], hence we do not applies an activation function on these three parameters. The range of the  $r_{u}$ are from 0 to $\infty$, the exponential function is used as activation. The range of the $\theta_{u}$, $\phi_{u}$ is from -1 to 1, hence a tanh activation function is applied. The Pointnet output before activation are denoted as [$x_{p}$, $y_{p}$, $z_{p}$, $r_{p}$, $\theta_{p}$, $\phi_{p}$]. Therefore, we have
\begin{equation}
\left\{
\begin{aligned}
& x_{u}, y_{u}, z_{u}=x_{p}, y_{p}, z_{p}, \\
& r_{u}=\exp(r_{p}), \\
& \theta_{u}, \phi_{u}= \tanh(\theta_{p}), \tanh(\phi_{p}). \\
\end{aligned}
\right.
\end{equation}
The output of the Pointnet can be remapped to their original value by following the description in Section \ref{section:method_estimation_annotation}.

\subsubsection{Network Training}
\begin{figure}[htbp]
	\centerline{\includegraphics[width=.45\textwidth]{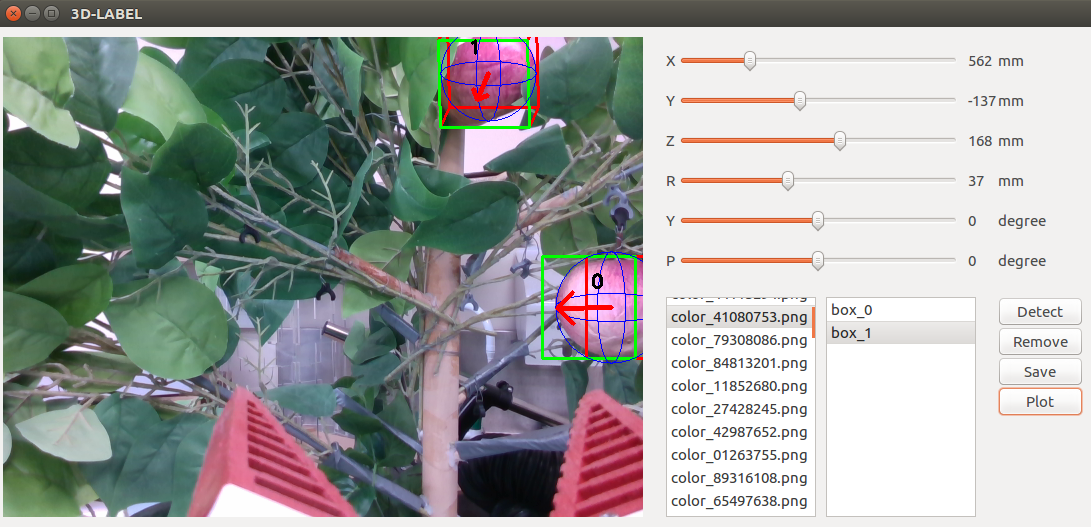}}
	\caption{The developed labelling tool for RGB-D images.}
	\label{fig:label_tool}
\end{figure}

The data labelling is performed on our own developed labelling tool, as shown in Figure \ref{fig:label_tool}. Our labelling tool records the six parameters of the 3D-OBB and all the points within the point clouds. The training of the Pointnet for 3D-OBB prediction is independent of the fruit recognition network training. There are 570 1-viewed point clouds of apples labelled in total (250 are collected in lab, 250 are collected in orchards). We apply 300 point sets as the training set (150 in each data set), 50 samples as validation set (25 in each data set), and the rest 220 samples as test set (110 in each data set). We introduce scaling (0.8 to 1.2), translation (-15 cm to 15 cm on each axis), rotation (-$10^{\circ}$ to $10^{\circ}$ on $\theta$ and $\phi$), adding Gaussian noise (mean equals 0, variance equals 2cm), and adding outliers (1\% to 5\% in total number of point clouds) in the data augmentation. One should notice that the orientation of samples after augmentation should still in the range between $-\frac{1}{4}\pi$ and $\frac{1}{4}\pi$.

The square error between prediction and ground truth is applied as the training loss. The Adam-optimizer in Tensorflow is used to perform the optimisation. The learning rate, decay rate, and total training epoch of the applied optimiser are 0.0001, 0.6 /epoch, and 100 epochs, respectively.

\subsection{Implementation Details}\label{section:method_implementation}
\subsubsection{System Configuration and Software}
The Intel-D435 RGB-D camera is applied in this research, a laptop (DELL-INSPIRATION) with Nvidia-GPU GTX-980M and Intel-CPU i7-6700 is used to control the RGB-D camera and perform the test. The connection between RGB-D camera and laptop is achieved by using the RealSense communication package in the Robot Operation System (ROS) in kinetic version \cite{ros} on the Linux Ubuntu 16.04. The calibration between the colour image and the depth image of the RGB-D camera is included in the realsense-ros. The implementation code of the Pointnet (in Tensorflow) is from the Github \cite{tensorflow-Pointnet}, and it is trained on the Nvidia-GPU GTX-980M. The implementation code of the Dasnet is achieved by using Tensorflow. The training of the Dasnet is performed on the Nvidia-GPU GTX-1080Ti. In the autonomous harvesting experiment, an industry robotic arm Universal Robot UR5 is applied. The communication between UR5 and the laptop is performed by using universal-robot-ROS. MoveIt! \cite{moveit} with TackIK inverse kinematic solver \cite{tracIK} is used in the motion planning of the robotic arm.

\subsubsection{Point Clouds Pre-processing}
An Euclidean distance based outlier rejection algorithm is applied to filter out outliers within point clouds before it is processed by Pointnet. When the distance between a point and point clouds centre is two times larger than the mean distance between the points and centre, we consider this point as an outlier and reject it. This step is repeated three times to ensure the efficiency of rejection. To improve the inference efficiency, a voxel downsampling function (resolution 3 mm) from the 3D data processing library open3D is used. Then we randomly pick 200 points from the downsampled point sets as the input of the Pointnet grasping estimation. The point set with the number of points less than 200 after voxel downsampling will be rejected since the insufficient number of points are presented.

\section{Experiment and Discussion}\label{section:experiment}
\subsection{Experiment Setup}
\begin{figure}[ht]
	\centering
	\includegraphics[width=.45\textwidth]{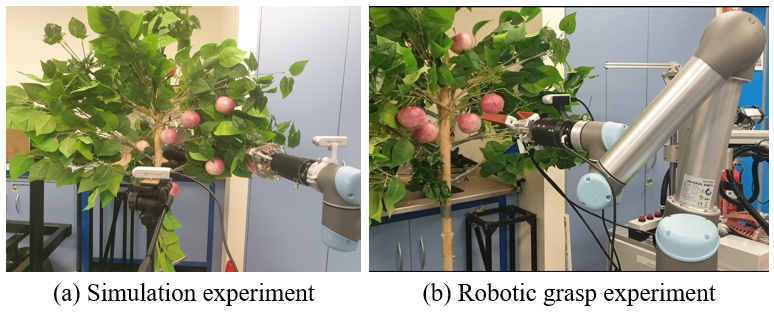}
	\caption{Experiment setup in laboratory scenario.}
	\label{fig:lab_test}
\end{figure}
We evaluate our proposed fruit recognition and grasping estimation algorithm in both simulation and the robotic hardware. In the simulation experiment, we perform the proposed method in the RGB-D data on the test set, which includes 110 point sets respectively in the laboratory environment and orchard environment. In the robotic harvesting experiment, we apply the proposed method to guide the robotic arm to perform the grasp of applies on the artificial plant in the lab. We apply IoU between predicted and ground-truth bounding box to evaluate the accuracy of 3D localisation and shape estimation of the fruits. We use 3D Axis Aligned Bounding Boxes (3D-AABB) to simplify the IoU calculation of 3D bounding box \cite{20193diou}. The IoU between 3D-AABB is denoted as IoU$_{3D}$. We set 0.75 (thres$_{IoU}$) as the threshold value for IoU$_{3D}$ to determine the accuracy of fruit shape prediction. In terms of the evaluation of the grasp pose estimation, we apply absolute error between the predicted value and ground truth value of grasp pose, as it can intuitively show the accuracy of predicted grasp pose. The maximum accepted error of grasp pose estimation for the robot to perform a successful grasp is 8$^{\circ}$, which is set as the threshold value in the grasp pose evaluation. This experiment is conducted in several scenarios, including noise and outlier presented conditions, and also dense clutter condition.

\subsection{Simulation Experiments}
In the simulation experiment, we compare our method with traditional shape fitting methods, which include sphere Random Sample Consensus (sphere-RANSAC) \cite{2007ransac} and sphere Hough Transform (sphere-HT) \cite{2007HT}, in terms of accuracy on fruit localisation and shape estimation. Both RANSAC and HT based algorithms take point clouds as input and generate the prediction of the fruit shape. The 3D bounding box of predicted shapes are then used to perform accuracy evaluation and compared with our method. This comparison are conducted on RGB-D images collected from both laboratory and orchard scenarios. 
\begin{figure}[ht]
	\centering
	\includegraphics[width=.45\textwidth]{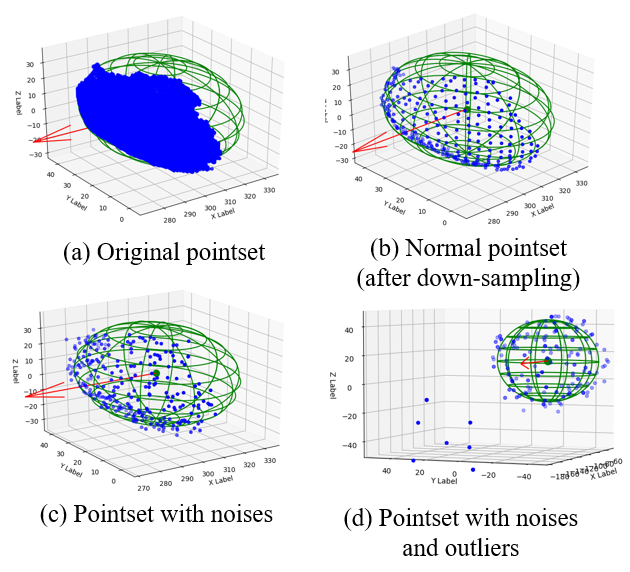}
	\caption{Pointset under different conditions, green sphere is the ground truth of the fruit shape.}
	\label{fig:pointset}
\end{figure}

\subsubsection{Experiments in laboratory Environments}
\begin{figure}[ht]
	\centering
	\includegraphics[width=.45\textwidth]{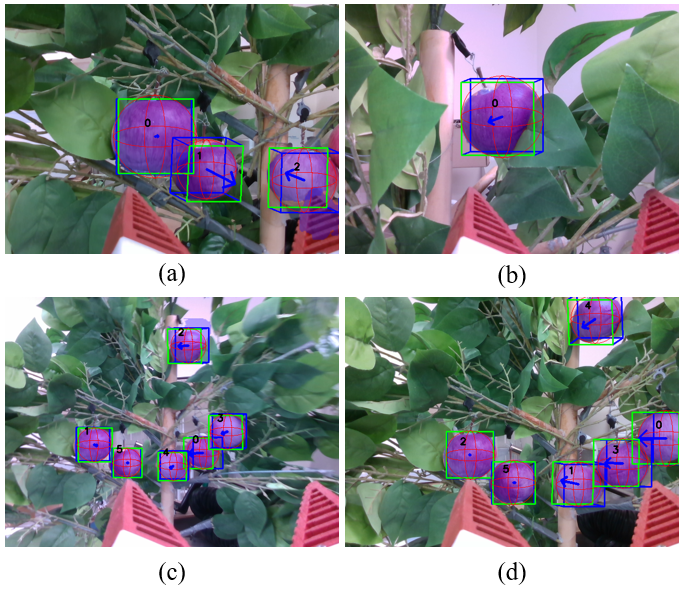}
	\caption{grasping estimation by using Pointnet. The green box are the front of the 3D-OBB, blue arrows are the predicted grasp pose, red sphere are the predicted shape of the fruits.}
	\label{fig:lab_modelling}
\end{figure}

\begin{table}[h] 
	\centering
	\caption{Accuracy of the fruit shape estimation by using Pointnet, RANSAC, and HT in different tests.}
	\begin{tabular}{c|cccccc}
		\hline
		&Normal&Noise&Outlier&Dense clutter&Combined\\
		\hline
		Pointnet&0.94&0.92&0.93&0.91&0.89\\
		RANSAC&0.82&0.71&0.81&0.74&0.61\\
		HT&0.81&0.67&0.79&0.73&0.63\\
		\hline
	\end{tabular}
	\label{table:table_1}
\end{table}
We performed Pointnet grasping estimation, RANSAC, and HT on the collected RGB-D images from the laboratory environment. The experimental results of three methods in different tests are shown in Table \ref{table:table_1}. From the experimental results, Pointnet grasping estimation significantly increases the localisation accuracy of the 3D bounding box of the fruits. Pointnet grasping estimation achieves 0.94 on IoU$_{3D}$, which is higher than the RANSAC and HT methods, respectively. To evaluate the robustness of different methods when dealing with noisy and outlier conditions, we randomly add Gaussian noise (mean equals 0, variance equals 2cm) and outlier (1\% to 5\% in the total number of point clouds) to the point clouds, as shown in Figure \ref{fig:pointset}. Three methods show similar robustness when dealing with outliers. Since both RANSAC and HT apply vote framework to estimate the primitives of the shape, which is robust to the outlier. However, when dealing with the noisy environment, Pointnet grasping estimation achieves better robustness, as compared to the RANSAC and HT. Since noisy point clouds can influence the accuracy of vote framework to a large extent. We also tested Pointnet grasping estimation, RANSAC, and HT in dense clutter condition. grasping estimation in dense clutter condition is challenging since the point clouds of objects can be influenced by other neighbouring objects. Pointnet grasping estimation can robustly perform accurate localisation and shape fitting of apples in this condition, which shows a significant improvement, as compared to the performance of the RANSAC and HT algorithms. The experimental results obtained by using Pointnet grasping estimation are presented in Figure \ref{fig:lab_modelling}, and the 3D-OBBs are projected into image space by using the method applied in the work of Novak \cite{2017vehicle}.

\begin{table}[h] 
	\centering
	\caption{Mean error of grasp orientation estimation by using Pointnet in different tests.}
	\begin{tabular}{c|cccccc}
		\hline
		&Normal&Noise&Outlier&Dense clutter&Combined\\
		\hline
		Pointnet&3.2$^{\circ}$&5.4$^{\circ}$&4.6$^{\circ}$&4.8$^{\circ}$&5.5$^{\circ}$\\
		\hline
	\end{tabular}
	\label{table:table_2}
\end{table}
In terms of the evaluation of the grasp orientation estimation, Pointnet grasping estimation shows accurate performance in the experimental results, as shown in Table \ref{table:table_2}. The mean error between predicted grasp pose and ground truth grasp pose is 3.2$^{\circ}$. Experimental results also show that Pointnet grasping estimation can accurately and robustly determine the grasp orientation of the objects in noisy, outlier presented, and dense clutter conditions. 

\subsubsection{Experiments in Orchards Environments}
\begin{figure}[ht]
	\centering
	\includegraphics[width=.45\textwidth]{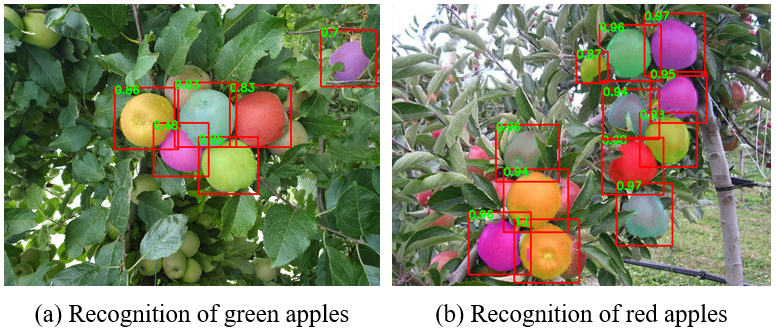}
	\caption{Detection and instance segmentation performed by using Dasnet on collected RGB images.}
	\label{fig:recognition_result}
\end{figure}

\begin{table}[h] 
	\centering
	\caption{Performance evaluation of fruit recognition in RGB-D images collected in orchard scenarios}
	\begin{tabular}{c|cccccc}
		\hline
		&F$_{1}$ score&Recall&Accuracy&IoU$_{mask}$\\
		\hline
		Dasnet&0.873&0.868&0.88&0.873\\
		\hline
	\end{tabular}
	\label{table:table_3}
\end{table}
In this experiment, we performed the fruit recognition (Dasnet) and Pointnet grasping estimation on the collected RGB-D images from apple orchards. The performance of the Dasnet is evaluated by using the RGB images in test set. We apply F$_{1}$ score and IoU as the evaluation metric of the fruit recognition. IoU$_{mask}$ stands the IoU value of instance mask of fruits in colour images. Table \ref{table:table_3} show the performance of the Dasnet (in terms of the detection accuracy and recall) and Pointnet grasping estimation, Figure \ref{fig:recognition_result} shows fruit recognition results by using Dasnet on test set. Experimental results show that Dasnet performs well on fruit recognition in orchard environment, having 0.88 and 0.868 on accuracy and recall, respectively. The accuracy of the instance segmentation on apples is 0.873. The inaccuracy of the fruit recognition is due to the illumination and fruit appearance variances. From the experiments, we found that Dasnet can accurately detect and segment the apples in the most conditions.

\begin{figure}[ht]
	\centering
	\includegraphics[width=.42\textwidth]{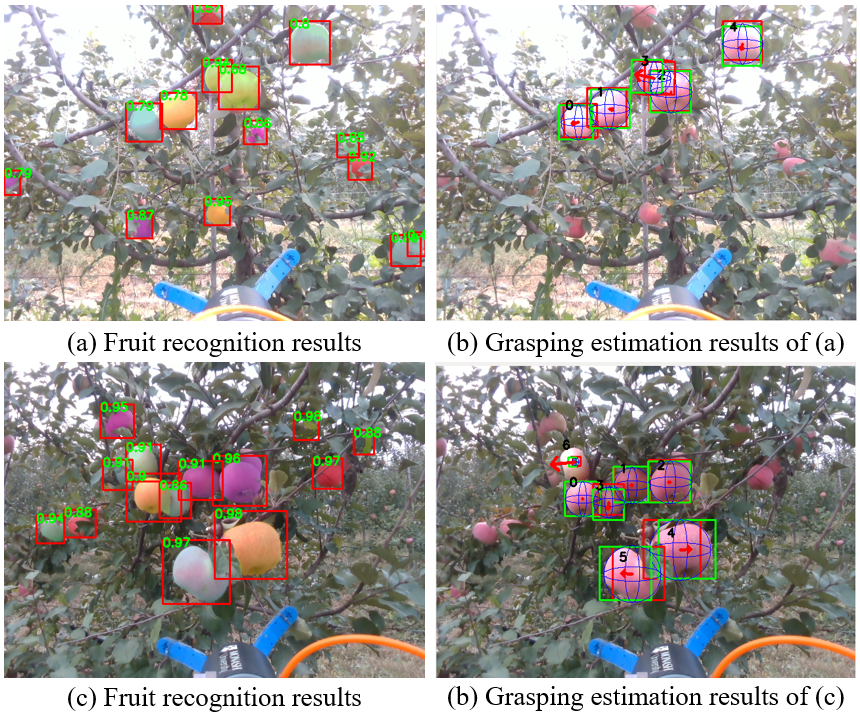}
	\caption{Fruit recognition and grasping estimation experiments in orchard scenario.}
	\label{fig:orchard_experiment}
\end{figure}

\begin{figure}[ht]
	\centering
	\includegraphics[width=.43\textwidth]{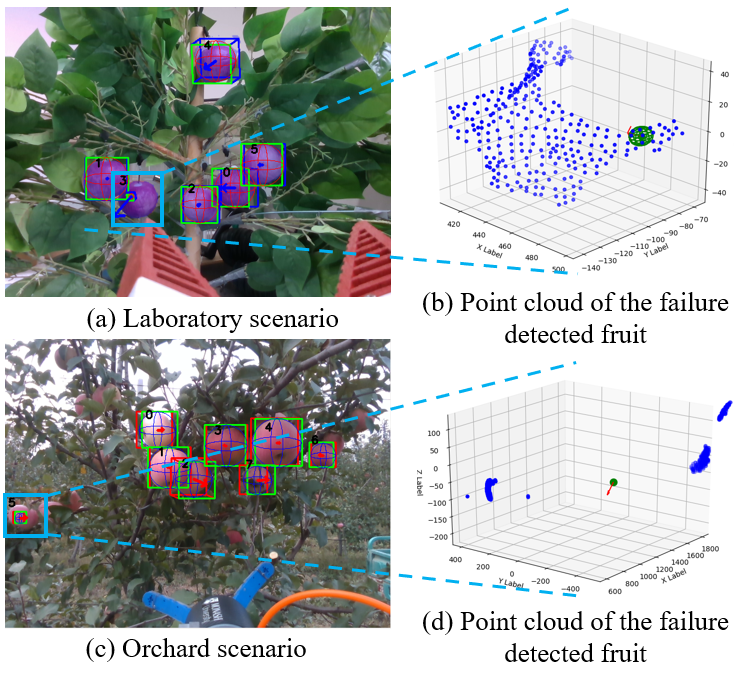}
	\caption{Failure grasping estimation in laboratory and orchard scenarios.}
	\label{fig:failure}
\end{figure}

\begin{table}[h] 
	\centering
	\caption{Evaluation on grasp pose estimation by using Pointnet in different tests in the orchard scenario.}
	\begin{tabular}{c|cccccc}
		\hline
		&Pointnet&RANSAC&HT\\
		\hline
		Accuracy&0.88&0.76&0.78\\
		Grasp Orientation&5.2$^{\circ}$&-&-\\
		\hline
	\end{tabular}
	\label{table:table_4}
\end{table}
Table \ref{table:table_4} shows the performance comparison between Pointnet grasping estimation, RANSAC, and HT. In the orchard environments, grasp pose estimation is more challenging compared to the indoor environments. The sensory depth data can be affected by the various environmental factors, as shown in Figure \ref{fig:failure}. In this condition, the performance of the RANSAC and HT show the significant decrease from the indoor experiment while Pointnet grasping estimation shows better robustness. The IoU$_{3D}$ achieved by Pointnet grasping estimation, RANSAC, and HT in orchard scenario are 0.88, 0.76, and 0.78, respectively. In terms of the grasp orientation estimation, Pointnet grasping estimations show robust performance in dealing with flawed sensory data. The mean error of orientation estimation by using Pointnet grasping estimation is 5.2$^{\circ}$, which is still within the accepted range of orientation error. The experimental results of grasp pose estimation by using Pointnet grasping estimation in orchard scenario is shown in Figure \ref{fig:orchard_experiment}.

\subsubsection{Common Failures in grasping estimation}
The major reason leading to the grasping estimation failure by using Pointnet grasping estimation is due to the sensory data defect, as shown in Figure \ref{fig:failure}. When under this condition, the results of Pointnet grasping estimation will always predicts a sphere with a very small value of radius. We can apply a radius value threshold to filter out this kind of failure during the operation. 

\subsection{Experiments of Robotic Harvesting}
\begin{figure}[ht]
	\centering
	\includegraphics[width=.47\textwidth]{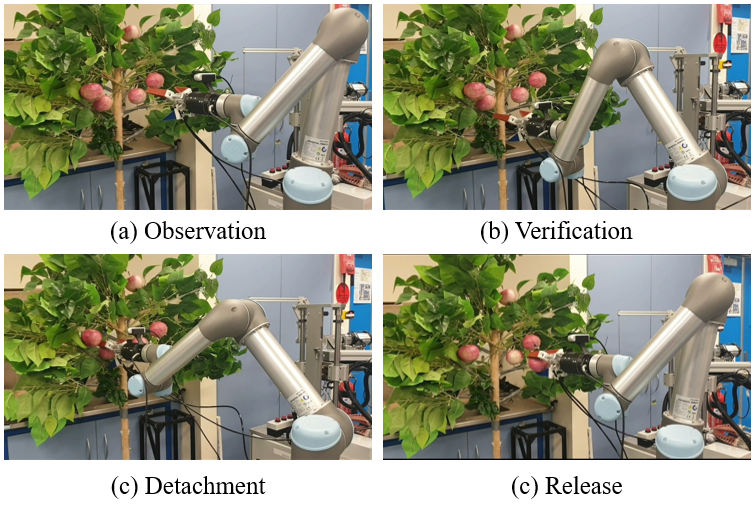}
	\caption{Autonomous harvesting experiment in the laboratory scenario.}
	\label{fig:robot_procedure}
\end{figure}
The Pointnet grasping estimation was tested by using a UR5 robotic arm to validate its performance in the real working scenario. We arranged apples on a fake plant in the laboratory environment, which is shown in Figure \ref{fig:lab_test}. We conducted multiple trails (each trail contains three to seven apples on the fake plant) to evaluate the success rate of the grasp. The success rate records a fraction of success grasps in the total number of grasp attempts. The operational procedures follow the design of our previous work \cite{2019visual}, as shown in Figure \ref{fig:robot_procedure}. We simulate the real outdoor environments of autonomous harvesting by adding noises and outliers into the depth data. We also tested our system in dense clutter condition. The experimental results are shown in Table \ref{table:table_5}. 
\begin{table}[h] 
	\centering
	\caption{Experimental results on robotic grasp by using Pointnet grasping estimation in Laboratory scenario}
	\begin{tabular}{c|cccccc}
		\hline
		&Normal&Noise&Outlier&Dense clutter&Combined\\
		\hline
		success rate&0.91&0.87&0.90&0.84&0.837\\
		\hline
	\end{tabular}
	\label{table:table_5}
\end{table}

From the experimental results presented in Tabla \ref{table:table_5}, Pointnet grasping estimation performs efficiently in the robotic grasp tests. Pointnet grasping estimation achieves accurate grasp results on normal, noise, and outlier conditions, which are 0.91, 0.87, and 0.9, respectively. In dense clutter condition, the success rate shows a decrease compared to the previous conditions. The reason for the success rate decreasing in dense clutter condition is due to the collision between gripper and fruits side by side. When collision presented in the grasp, it will cause the shift of the target fruit and lead to the failure of the grasp. This defect can be either improved by re-design the gripper or propose multiple grasp candidates to avoid the collision. The collision between gripper and branches can also lead to grasping failure in the other three conditions. Although such defect can affect the success rate of robotic grasp, it still achieves good performance in experiments. The success rate of robotic grasp under dense clutter condition and that all of factors combined condition are 0.84 and 0.837, respectively. The average running time of the fruit recognition and grasping estimation for one frame RGB-D image (5-7 apples included) is about 0.32 seconds on GTX-980M, showing a real-time ability to be performed in the robotic harvesting.

\section{Conclusion and Future Work}\label{section:conclusion} 
In this work, a fully deep-learning neural network based fruit recognition and grasping estimation method were proposed and validated. The proposed method includes a multi-functional network for fruit detection and instance segmentation, and a Pointnet grasping estimation to determine the proper grasp pose of each fruit. The proposed multi-function fruit recognition network and Pointnet grasping estimation network was validated in RGB-D images taken from the laboratory and orchard environments. Experimental results showed that the proposed method could accurately perform visual perception and grasp pose estimation. The Pointnet grasping estimation was also tested with a robotic arm in a controlled environment, which achieved a high grasping success rate (0.847 in all factor combined condition). Future works will focus on optimising the design of the end-effector and validating the developed robotic system in the coming harvest season.

\section*{Acknowledgement}
This research is supported by ARC ITRH IH150100006 and THOR TECH PTY LTD. We also acknowledge Zhuo Chen for her assistance in preparation of this work.

\bibliographystyle{ieeetr}
\bibliography{reference}
\renewcommand\refname{References}   
\end{document}